%% file: acl_2026_text-simp.tex
\documentclass[11pt]{article}

\usepackage[]{acl}

\usepackage{times}
\usepackage{latexsym}

\usepackage[T1]{fontenc}

\usepackage[utf8]{inputenc}

\usepackage{microtype}

\usepackage{inconsolata}

\usepackage{graphicx}

\input{00-commonz}

\newcommand{\GITPage}{https://github.com/bilha-analytics/text-simplification-metrics-eval/tree/main/analysis_output}

\newcommand{\TheTitle}{
	Making Knowledge Accessible: Divergent Readability-Accuracy Strategies of Mistral and QWen in Biomedical Text Simplification}

\title{\TheTitle}

\author{P. Bilha Githinji \\\And
  Aikaterini Melliou \\\And  
  Zeming Liang \\\And  
  Lian Zhang \\\And  
  Peiwu Qin
  \\}

\begin{document}
\maketitle
\begin{abstract}
	\input{\CHAPTERSDIR/00-abstract}
\end{abstract}

 
\section{Introduction}
\label{sec:dpintro}
\input{\CHAPTERSDIR/01-a-introduction}

\section{Method}
\label{sec:dpmethod}
\input{\CHAPTERSDIR/02-a-methodology}

\section{Results}
\label{sec:dresults}

\input{\CHAPTERSDIR/03-results}

\section{Discussion}
\label{sec:ddiscuss}
\input{\CHAPTERSDIR/04-a-discussion}

\section{Conclusion}
\label{sec:dconc} 
\input{\CHAPTERSDIR/05-a-conclusion}


\section*{Limitations}

\input{\CHAPTERSDIR/04-b-limitations}

\section*{Acknowledgments}



{
	\bibliography{\dpaperbib}
}

\appendix


\input{\CHAPTERSDIR/99-appendix}

\end{document}

%% file: 00-commonz.tex
\usepackage{amsmath}
\usepackage{amssymb}

\usepackage[utf8]{inputenc} 
\usepackage[T1]{fontenc}    
\usepackage{hyperref}       
\usepackage{url}            
\usepackage{booktabs}       
\usepackage{amsfonts}       
\usepackage{nicefrac}       
\usepackage{microtype}      
\usepackage{cleveref}       
\usepackage{graphicx}

\usepackage{booktabs, threeparttable} 
\usepackage{longtable}
\usepackage{tabularx}
\usepackage[justification=centerlast]{caption} 
\usepackage{subcaption} 
\usepackage{xcolor}

\usepackage{relsize}  

\usepackage{multicol}

\usepackage{listings} 

\definecolor{codegreen}{rgb}{0,0.6,0}
\definecolor{codegray}{rgb}{0.5,0.5,0.5}
\definecolor{codepurple}{rgb}{0.58,0,0.82}
\definecolor{backcolour}{rgb}{0.95,0.95,0.92}

\lstdefinestyle{promptstyle}{
	backgroundcolor=\color{backcolour},   
	commentstyle=\color{codegreen},
	keywordstyle=\color{magenta},
	numberstyle=\tiny\color{codegray},
	stringstyle=\color{codepurple},
	basicstyle=\ttfamily\footnotesize,
	breakatwhitespace=false,         
	breaklines=true,                 
	captionpos=b,                    
	keepspaces=true,                 
	numbers=left,                    
	numbersep=5pt,                  
	showspaces=false,                
	showstringspaces=false,
	showtabs=false,                  
	tabsize=2,
	frame=single,
	rulecolor=\color{black}
}

\definecolor{codegreen}{rgb}{0,0.6,0}
\definecolor{codegray}{rgb}{0.5,0.5,0.5}
\definecolor{codepurple}{rgb}{0.58,0,0.82}
\definecolor{backcolour}{rgb}{0.95,0.95,0.92}

\lstdefinestyle{pythonstyle}{
	backgroundcolor=\color{backcolour},   
	commentstyle=\color{codegreen},
	keywordstyle=\color{magenta},
	numberstyle=\tiny\color{codegray},
	stringstyle=\color{codepurple},
	basicstyle=\ttfamily\footnotesize,
	breakatwhitespace=false,         
	breaklines=true,                 
	captionpos=b,                    
	keepspaces=true,                 
	numbers=left,                    
	numbersep=5pt,                  
	showspaces=false,                
	showstringspaces=false,
	showtabs=false,                  
	tabsize=2,
	frame=single,
	rulecolor=\color{black},
	language=Python
}
 
\crefname{section}{Sec.}{Secs.}
\Crefname{section}{Section}{Sections}
\Crefname{table}{Table}{Tables}
\crefname{table}{Tab.}{Tabs.}

\newcommand{\dpaperbib}{t1-plabz-citations}

\newcommand*{\zsubsection}{\subsection} 
 
\newcommand*{\boldtextsection}{\paragraph} 
\newcommand*{\italictextsection}{\textbf} 

\newcommand{\NMETRICZ}{21}

\newcommand{\MISTRAL}{Mistral-Small 3 24B}
\newcommand{\QWEN}{Qwen 2.5 32B}
 
\newcommand{\lblMISTRALflexi}{Mistral - flexi}
\newcommand{\lblMISTRALstrict}{Mistral - strict}
\newcommand{\lblQWENflexi}{QWen - flexi}
\newcommand{\lblQWENstrict}{QWen - strict}

%% file: 00-abstract.tex
The growing public demand for accessible biomedical information calls for scalable text simplification. 
While large language models (LLMs) offer solutions, they too struggle with balancing improved readability against preservation of meaning. 
This report empirically compares how two LLMs - instruction-tuned~\MISTRAL~and the reasoning-augmented~\QWEN~- navigate this trade-off in biomedical text simplification, benchmarked against human performance.
Our analysis highlights how each model applies distinct operational strategies when simplifying biomedical text. 
Mistral exhibits a tempered lexical simplification approach that consistently enhances readability across multiple metrics while preserving discourse fidelity (BERTScore: $0.91$, statistically comparable to that of humans).
In comparison, QWen also attains enhanced readability performance and a reasonable BERTScore of $0.89$, but presents a disconnect in balancing between readability and accuracy. 
Additionally, a comprehensive correlation analysis of a suite of~\NMETRICZ~metrics confirms strong functional redundancies in metrics and informs adaptation requirements.

%% file: 01-a-introduction.tex
Access to understandable health information is fundamental to informed decision-making and public health~\citep{stacey_dawn_shared_2017,schlacher_guide_2024}.
Yet, patient-facing materials frequently exceed the recommended reading levels~\citep{stacey_dawn_shared_2017,mishra_comparison_2020,schlacher_guide_2024}, and the digital proliferation of biomedical content poses significant risks, including misinformation, oversimplification, and lack of necessary clinical context~\citep{suarez-lledo_prevalence_2021,patel_online_2024}.
Reliable automated text simplification offers a scalable solution to render complex clinical and scientific texts into plain language at scale, bridging the gap between expert knowledge and public understanding. 

Large language models (LLMs), trained on vast and varied corpora, possess inherent linguistic capabilities that may extend to text simplification out of the box.
Text simplification entails reducing specialized vocabulary and complex sentence structures while strictly maintaining semantic equivalence between source and output text~\citep{xu_optimizing_2016}. 
It is distinct from lay summarization, which introduces an additional content distillation step that requires balancing semantic equivalence with conciseness~\citep{khan_exploring_2023}.

Empirical evidence consistently demonstrates a persistent and critical performance trade-off between readability gains and content accuracy, with solutions, including LLM-based methods, often achieving high readability at the cost of factual inaccuracies, semantic drift, and undesirable omissions~\citep{maynez_faithfulness_2020,ondov_lessons_2025,wu_-depth_2025,agrawal_evaluation_2025}.  
Some studies emphasize that domain adaptation for biomedical text is necessary since LLMs do not reason over biomedical semantics to appropriately translate lexical and syntactic changes~\citep{wu_large_2024,demartini_llms_2025}.
Domain adaptation strategies for text simplification, however, report conflicting results, and some fail to outperform their general-purpose counterparts~\citep{feng_evaluation_2024,shao_-context_2024,balde_medvoc:_2024,yang_unveiling_2024,swanson_biomedical_2024,alamleh_readability_2025,ondov_lessons_2025,househ_plain_2025,househ_readability_2025,alamleh_readability_2025}. 
As for general-purpose LLMs, large architectural models such as GPT-4 and Llama3 70B exhibit superior performance or minimal numerical differences from domain-adapted alternatives, while small models underperform considerably~\citep{feng_evaluation_2024,dorfner_biomedical_2024,chen_benchmarking_2025}.

As LLMs become integrated into everyday information-seeking practices, it is increasingly important to understand their capacity to consistently navigate the tension between maximizing content readability and ensuring discourse fidelity and safety without fine-tuning or technical adaptations typically inaccessible to lay users. 
Moreover, rigorous assessment of automated text simplification requires consideration of both readability and preservation of semantic fidelity. 
While evidence establishes various readability formulas and cautions against traditional accuracy metrics~\citep{xu_optimizing_2016,agrawal_evaluation_2025,househ_plain_2025}, a comprehensive view of the associations within and between these two functional groups of metrics needs clarification.

Using biomedical abstracts, inherently compressed summaries with concentrated technical jargon, complex sentence structures, and high informational density, this study isolates and
comprehensively investigates the readability-accuracy trade-off of two medium-sized general-purpose LLMs that are a practical sweet-spot for research and practice. 
We comparatively investigate the instruction-tuned~\MISTRAL~\citep{noauthor_mistral_nodate} and the reasoning-augmented~\QWEN~\citep{qwen2.5} under two temperature configurations each, and benchmark their performance against that of human experts. 
Specific contributions include 
\begin{itemize}
	\item{Empirical assessment with only prompting, updating practical baselines and pointing to lexical simplification, as opposed to syntactic structure, as the main hurdle.
	}
	
	\item{Identification of an architectural advantage in the instruction-tuned Mistral model that is superior in readability, comparable to human experts with regard to discourse fidelity, and robust to temperature adjustment. 
	}
	
	\item{A rigorous assessment of the associations within and across readability and accuracy metrics, revealing how the readability-accuracy tension presents.
	} 
\end{itemize}

%% file: 02-a-methodology.tex

\zsubsection{Data}  
The primary benchmark is a public dataset of $750$ biomedical abstracts paired with human-simplified texts~\citep{attal_dataset_2023}. 
We refer to the process of text simplification by human experts as the human model, while the original scientific abstracts of this dataset serve as a curated control cohort (the control set) spanning 75 biomedical topics. 
An uncurated custom dataset is derived from a random sample of domain-specific abstracts covering Traditional Chinese Medicine (TCM) and Oncology. 
The selection of TCM and Oncology is partly motivated by the density of unique terminology in this subdomain and partly by the translational relevance given growing public interest in TCM for cancer symptom management~\citep{schuerger_evaluating_2019,trubner_health_2025}.

\zsubsection{Text simplification systems} 
We consider two architectural classes represented by~\MISTRAL~ model, an instruction-tuned model optimized for task fidelity, and~\QWEN, a reasoning-augmented model designed for complex problem solving. 
To assess robustness, we configure each with two temperature settings, namely: a $strict$ configuration with temperature $T=0.2$, and a relatively higher stochasticity state (tagged with $flexi$ for flexible) with $T=0.4$. 
The four resulting LLM simplification processes and the human-expert simplification process constitute the five plain-text adaptation systems in our study.

\zsubsection{Prompt design}
We specify the simplification process to the LLMs via a standardized, empirically developed, zero-shot prompt aimed at consistent reception of the task across models. 
The prompt defines the task as a direct sentence-by-sentence adaptation of the input text, explicitly prohibiting summarisation, and addressing both linguistic complexity (e.g., jargon replacement, splitting complex sentences) and discourse complexity (e.g., adding explanations, abstracting esoteric details).
The prompt design follows domain-agnostic established principles for plain language communication~\citep{cramm_best_2017,attal_dataset_2023}, and aligns with the human-benchmark process.  

Moreover, we incorporate a self-reporting mechanism where LLMs tag each output sentence with the applied simplification transform and associated rationale. 
Overall, the prompt design results in a controlled instrument for the assessment. The full prompt is presented in~\Cref{sec:appendix-prompt}.

\input{\FIGURESDIR/fig-mu-distz-all-min}

\zsubsection{Evaluation metrics and analysis}
We assemble a suite of commonly employed metrics for readability, discourse fidelity, and content safety. Foundational distributional metrics such as average sentence length and proportion of difficult words, inherent in readability formulas, are also tracked to provide fine-grained visibility into drivers of score variation. 
The metrics and their operating properties are detailed in~\Cref{sec:appendix-methods} and summarized next.

\textbf{Readability and coherence metrics.} 
This group comprises established readability formulas that quantify lexical and syntactic complexity, differing primarily in their treatment of word difficulty. 
Word difficulty is typically determined by polysyllabic counts, with notable exceptions being the Automated Readability Index (ARI), which relies on character count, and the Dale-Chall formula, which employs a predefined lexicon of 3000 words familiar to a U.S. fourth-grader.
Dale-Chall score, Gunning Fog Index, Flesch-Kincaid Grade Level (FKGL), SMOG Index, and ARI report performance as an estimated U.S. school grade level, where a lower score indicates improved simplification. 
SARI, a metric specifically formulated for the evaluation of automated text simplification processes, assesses the overall goodness against human reference.

\textbf{Content and discourse preservation metrics.} 
This group assesses semantic integrity, completeness, and subject matter relevance. 
Robust meaning preservation is evaluated using semantic congruence metrics, BERTScore, and document-level semantic similarity via LLM embeddings.  
Thematic consistency employs Latent Dirichlet Allocation (LDA) for topic modeling and vocabulary matching to trace handling of specialized jargon. 
While LDA is not ideal for short documents, abstracts have high information density and reasonable length for meaningful word co-occurrence.

\textbf{Content safety metrics.} 
We include safety checks using an established toxicity classifier, and recognize that, as a general-purpose tool, it may not capture safety risks such as oversimplification or misplaced emphasis. This inclusion hopes to foster continual monitoring necessary for the responsible integration of AI. 

For each metric, performance estimators are calculated as the document-level mean scores, while statistical comparisons employ Welch's t-test. 
All statistical results are reported at a significance level of $\alpha = 0.05$.

%% file: fig-mu-distz-all-min.tex
 
\begin{figure*}
    \centering  
\begin{subfigure}[b]{0.245\textwidth}
        \centering \raisebox{-\height}{\includegraphics[width=\linewidth, height=4cm]{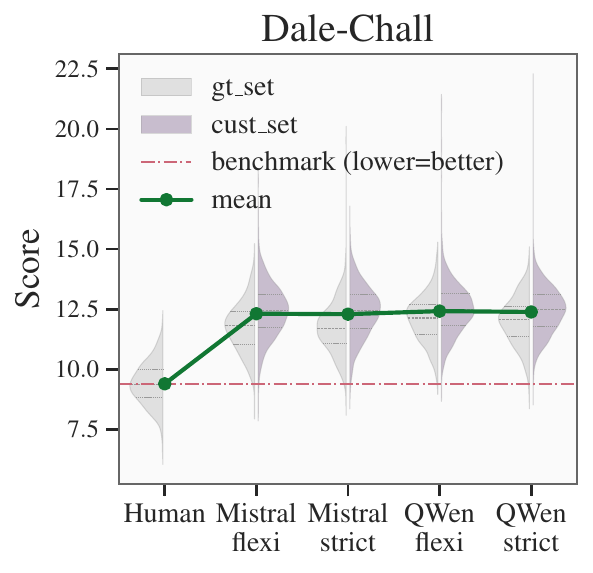}}
\label{fig:mean-distz-readability-dale-chall}
        \end{subfigure}
\begin{subfigure}[b]{0.245\textwidth}
        \centering \raisebox{-\height}{\includegraphics[width=\linewidth, height=4cm]{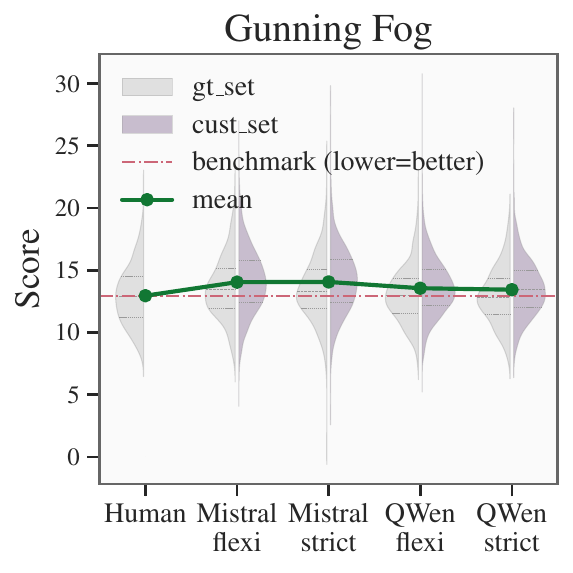}}
\label{fig:mean-distz-gunning fog}
        \end{subfigure}
\begin{subfigure}[b]{0.245\textwidth}
        \centering \raisebox{-\height}{\includegraphics[width=\linewidth, height=4cm]{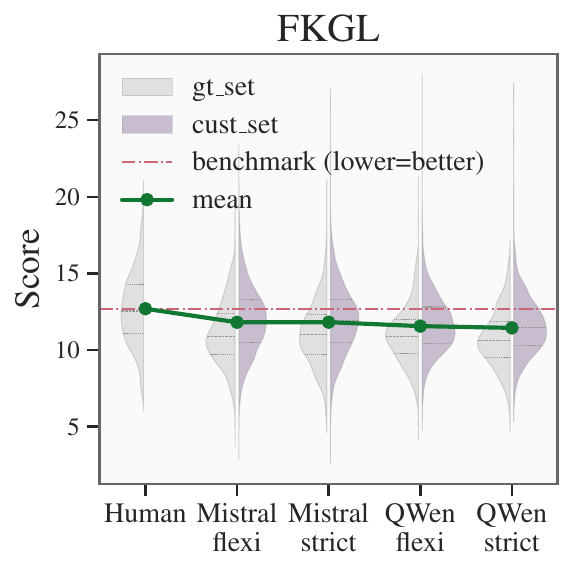}}
\label{fig:mean-distz-readability-flesch-grade}
        \end{subfigure}
\begin{subfigure}[b]{0.245\textwidth}
        \centering \raisebox{-\height}{\includegraphics[width=\linewidth, height=4cm]{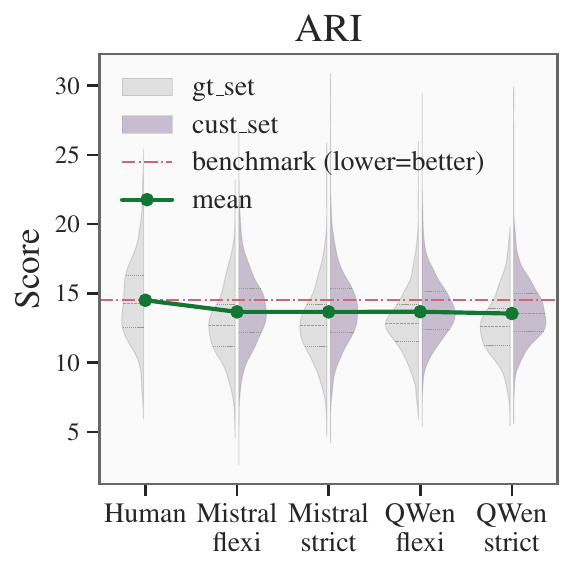}}
\label{fig:mean-distz-readability-ari}
        \end{subfigure}
\hfill\begin{subfigure}[b]{0.245\textwidth}
        \centering \raisebox{-\height}{\includegraphics[width=\linewidth, height=4cm]{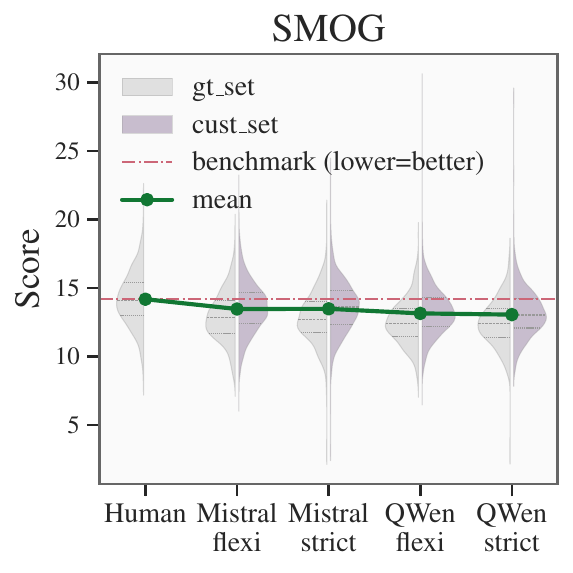}}
\label{fig:mean-distz-SMOG Index}
        \end{subfigure}
\begin{subfigure}[b]{0.245\textwidth}
        \centering \raisebox{-\height}{\includegraphics[width=\linewidth, height=4cm]{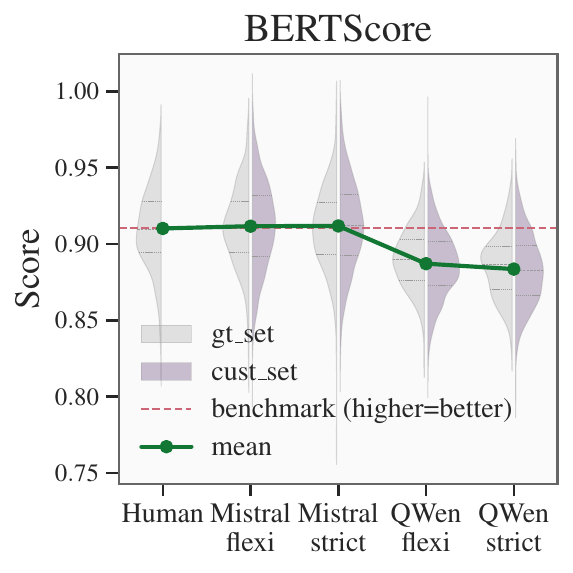}}
\label{fig:mean-distz-BERTScore}
        \end{subfigure}
\begin{subfigure}[b]{0.245\textwidth}
        \centering \raisebox{-\height}{\includegraphics[width=\linewidth, height=4cm]{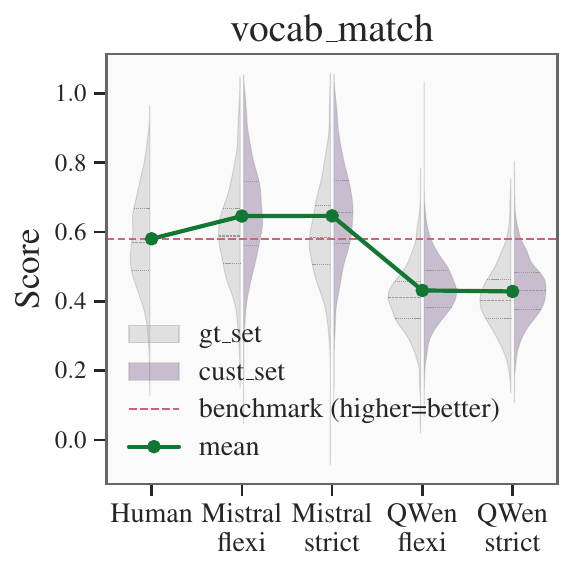}}
\label{fig:mean-distz-vocab-match}
        \end{subfigure}
\begin{subfigure}[b]{0.245\textwidth}
        \centering \raisebox{-\height}{\includegraphics[width=\linewidth, height=4cm]{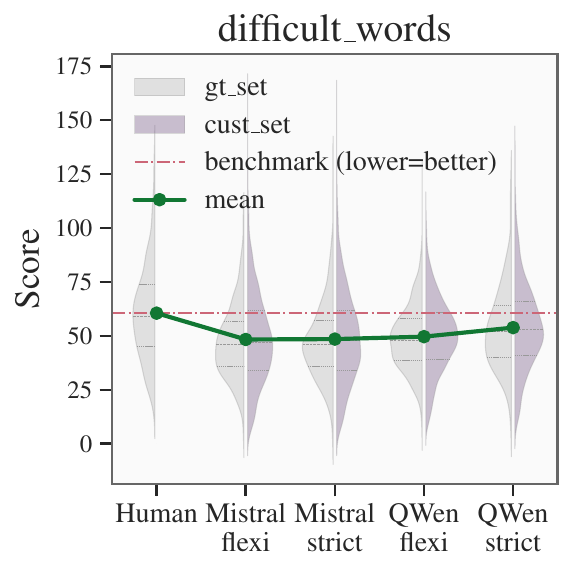}}
\label{fig:mean-distz-difficult words}
        \end{subfigure}
        
    \caption{Average performance and underlying data distributions}
    \label{fig:means-distz-mini}
    \end{figure*}

%% file: 03-results.tex


\input{\FIGURESDIR/fig-pvalz}

\zsubsection{System goodness}
The experimental groups and LLM completion rates (detailed in~\Cref{sec:appendix-methods}) yield sufficient samples for robust validation. 
System goodness is assessed via SARI~\citep{xu_optimizing_2016}, which compares LLM simplifications against human reference. 
As shown in~\Cref{tbl:sari}, both models advance over previous baselines (SARI 34, best-reported baseline for encoder-decoder models T5 and BART)~\citep{attal_dataset_2023}. 
Mistral outperforms QWen, achieving SARI scores of 42.46 (flexible, 95\% CI: 41.86 - 43.05) and 42.37 (strict, 95\% CI: 41.77 - 42.96), while QWen scores 38.38 (strict, 95\% CI: 38.28 - 38.47) and 37.84 (strict, 95\% CI: 37.16 - 38.52), indicating Mistral's superior simplification across both temperature settings.
Moreover, Mistral's performance approaches that of GPT-4.1-mini (best of five runs SARI score of 43.83)~\citep{ondov_lessons_2025}.

\input{\TABLESDIR/pv-sari}

\zsubsection{Readability and coherence}

Standard readability formulas are employed, and their mean results across the five simplification systems are presented in~\Cref{fig:means-distz-mini}, with human benchmark indicated as a threshold and associated statistical comparisons in~\Cref{fig:pvalz-grid-all}.

The LLMs achieve statistically superior readability on four of six metrics but underperform on Dale-Chall and Gunning Fog. 
This divergence reflects formulaic differences, where these two metrics particularly employ a predefined lexicon of familiar words. 
Dale-Chall index identifies the human benchmark as having the best readability with a U.S. school grade score of $9.40$ compared to $12.30$ and $12.39$ for Mistral and QWen, respectively. 
Conversely, Flesch-Kincaid Grade Level (QWen $11.45$, Mistral $11.81$, human $12.70$) and Flesch Reading Ease (QWen $43.83$, Mistral $42.11$, human $39.77$, higher is better) rank QWen best. 

While there are statistical differences between model scores, the numerical differences are small. In particular, collectively, the models have readability U.S. school grade scores in the range $12 - 14$ compared to the human benchmark's grade $9 - 15$, with the Dale-Chall index appearing most difficult for the LLMs. 
Temperature effects are similarly minimal, though Mistral demonstrates statistically consistent performance across both settings.

\input{\FIGURESDIR/fig-corr-matrix}


\zsubsection{Accuracy}
We assess accuracy through semantic congruence (BERTScore, semantic similarity score), topical relevance (LDA-topics score), traditional content preservation metrics (ROUGE-L, SacreBLUE), and underlying measures (vocabulary matching, difficult words proportion).
Full distributions and statistical results appear in~\Cref{fig:means-distz-mini,fig:pvalz-grid-all}.

On BERTScore and LDA-topics score, Mistral achieves human-level performance ($0.91$ and $0.37$, respectively) across temperature settings, with no statistical difference from the human benchmark.
QWen attains lower scores that vary with temperature, peaking under flexible configuration ($0.89$ BERTScore, $0.23$ LDA-topics score).  

Both models significantly reduce difficult words compared to human experts (Mistral flexi: $48.35\%$; QWen flexi: $49.66\%$; human: $60.55\%$). 
However, Mistral has the highest vocabulary retention and QWen the lowest (Mistral $0.65$, QWen $0.43$, human $0.58$), suggesting conservative treatment of specialized vocabulary by Mistral and exploration by QWen.
Mistral appears to have a conservative strategy that selectively balances lexical simplification with semantic fidelity. Conversely, QWen's approach, while achieving readability gains (presented previously), risks greater semantic displacement. 
We revisit this trade-off under correlation analysis.



\zsubsection{Content Safety}
The results in~\Cref{fig:pvalz-grid-all} show that all five plain-text adaptation processes achieve mean toxicity scores of virtually zero, and that LLM outputs are not statistically distinguishable from human-simplified text with regards to toxicity. 
This outcome, while not surprising given the benign nature of the source biomedical abstracts, provides necessary empirical validation of safety constraints and does not obviate the need for continual monitoring.


\zsubsection{Associations between metrics}
Here, we consider the tension between readability and accuracy by analysing inter-metric relationships. 
Pairwise correlation matrices ($\alpha = 0.05$) are visualized in~\Cref{fig:corr-matrix-grid}, with representative distributional pair-plots provided in~\Cref{sec:appendix-results}.

\italictextsection{Readability and coherence.} 
Readability metrics exhibit strong functional congruence and high redundancy within the metric set. Five of six indicators show statistically significant correlations $\ge 0.7$  across human and both LLM architectures (accounting for the directionality of the Flesch Reading Ease score).
Dale-Chall correlates more weakly (coefficients in the range $[0.4, 0.6]$), reiterating its relative difficulty for the LLMs. 

All systems (human and LLMs) appear to prioritise syntactic over lexical simplification. Correlations for sentence length (average words per sentence) are in the range $[0.4, 0.7]$, while difficult words correlate weakly (coefficients range $[0.1, 0.4]$). 
This suggests that lexical control, rather than syntactic restructuring, is the primary hurdle, assuming the formulas are suitably calibrated. 

Moreover, architectural differences emerge in lexical treatment. 
Correlations between readability formulas and difficult words are highest for Mistral compared to Qwen (human $0.2-0.3$, Mistral $0.4$, Qwen $0.1$), whose lexical scores present as less connected to readability formulas despite competitive overall scores.

Additionally, the LLMs appear to achieve readability gains more efficiently, since their compression ratios are below 1 (document length decreases), while humans expand the text during simplification. Mistral and QWen address readability with greater conciseness, a practical advantage for scalable simplification.


\italictextsection{Accuracy.} 
Similarly, discourse preservation metrics exhibit internal coherence, with BERTScore appearing to encompass the signal embedded in traditional lexical metrics (ROUGE-L and SacreBLEU), as correlation coefficients are $\ge 0.9$.   
Furthermore, while humans use text expansion as a readability-enhancing tactic, the LLMs appear to primarily expand text for the benefit of semantic integrity. 
The correlation between BERTScore and words compression ratio is substantially stronger for LLM outputs ($~ 0.6$) than for human simplification ($~ 0.2$).

\input{\FIGURESDIR/fig-changes-min}

\boldtextsection{Cross-function associations.} 
Cross-functional analysis reveals divergent algorithmic strategies between instruction-tuned Mistral and reasoning-augmented QWen. 
Associations between readability and accuracy, while weakly correlated, have coefficient magnitudes that are larger for Mistral (coefficients range $[0.2, 0.4]$) than for QWen ($[-0.2, 0.1]$), suggesting Mistral yields output that better aligns accuracy and readability optimisation. 
Similarly, Mistral has larger correlation coefficients for accuracy and word difficulty (coefficients in the range $[0.2, 0.5]$ for Mistral, $[-0.3, 0.0]$ for QWen, and $[0.1, 0.3]$ for human experts). This association is relatively stronger ($0.5$) for traditional lexical alignment accuracy metrics (ROUGE-L, SacreBLEU), pointing to Mistral's conservative retention-oriented treatment of specialised vocabulary. As for BERTScore, however, correlation with word difficulty is weaker and positive for Mistral ($0.2$) and human experts ($0.1$) but weakly negative for QWen ($-0.3$), reiterating QWen's treatment of lexical complexity that risks semantic integrity.

Regression on principal components of the metrics reinforces these findings. 
Human simplification yields the highest adjusted $R^2$ for both readability (Dale-Chall: $0.39$; Flesch: $0.45$) and accuracy (BERTScore: $0.49$),
followed by Mistral (Dale-Chall: $0.34$; Flesch: $0.37$; BERTScore: $0.28$), and then QWen (Dale-Chall: $0.26$; Flesch: $0.23$; BERTScore: $0.29$). 
Humans appear to jointly optimise readability and accuracy objectives, while LLMs show unidirectional responsiveness, where readability improves with accuracy, but accuracy does not correspondingly respond to measured readability aspects. Additionally, Mistral's tempered lexical control is further reflected in higher $R^2$ for difficult words and vocabulary matching (details in~\Cref{sec:appendix-methods}).

It is noteworthy, though, that by itself, QWen is still a capable model for the simplification task; its aggressive lexical substitution yields superior readability (by most of the formulas) and a reasonable BERTScore (0.89). Its weaker associations may reflect exploration of its broader lexical search space, warranting further investigation. 
These results may also signal limitations in capturing LLM simplification processes with the current suite of metrics.



\zsubsection{Self-reported rationale on changes made.} 
Analysis of the self-reported rationales offers additional insights into each model's simplification style, quantifying choices made against standardised transforms detailed under prompt design in~\Cref{sec:appendix-prompt} and summarised in~\Cref{fig:changes-pairs-mini}.  

Both models rely on jargon/parlance swapping as their dominant simplification strategy but exhibit divergent secondary tactics. 
Mistral favours omitting superfluous details or retaining original text, treating complex phrases or vocabulary cautiously. 
Conversely, QWen engages in conceptual expansion, adding explanatory context or abstracting complex concepts. 
Underperforming samples (observations with below the average benchmark scores) receive more of these secondary tactics in both models. 
These patterns reflect underlying design philosophies of the models.  
Mistral is conservative and operates primarily within the bounds of the input, while QWen engages in conceptual exploration. For Mistral, this constrained behaviour seems to favour tempered simplification that preserves discourse fidelity.

%% file: fig-pvalz.tex
 
\begin{figure*}[htbp!]  
    \centering  
    \includegraphics[width=1.\textwidth]{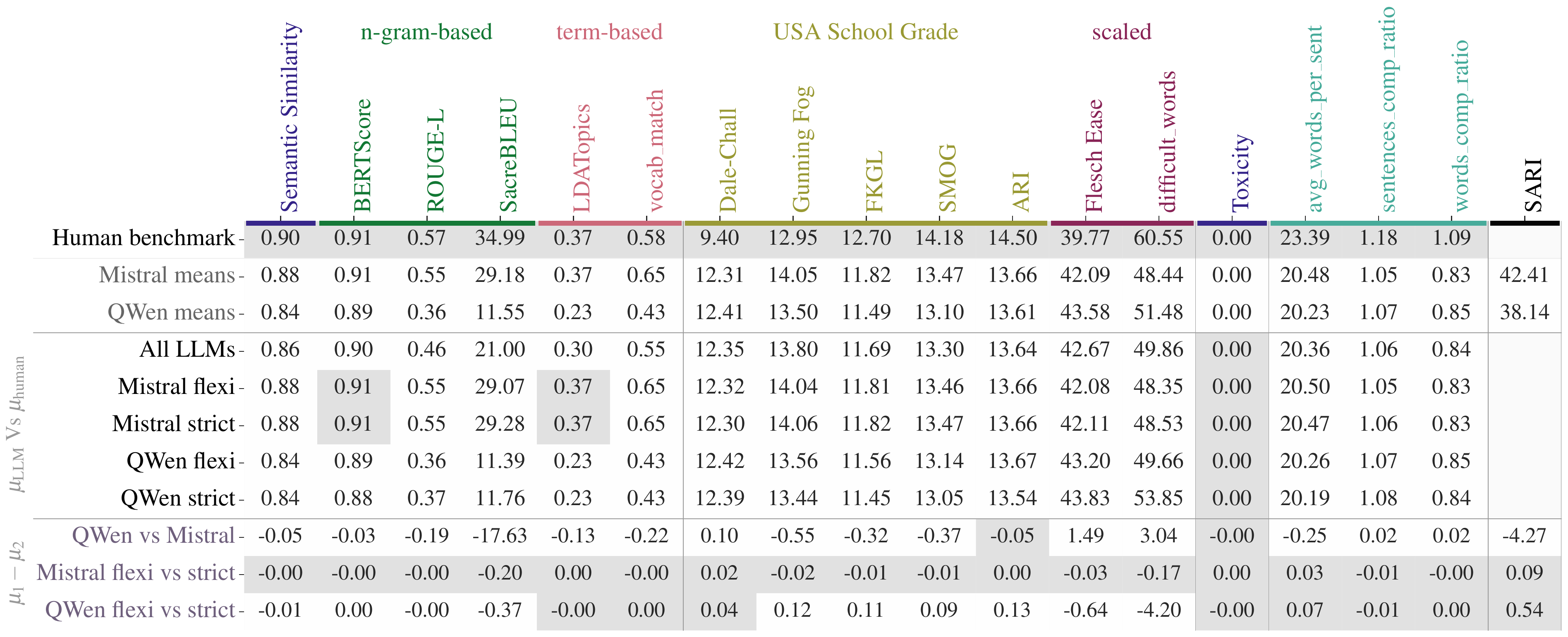}
    \caption{Hypotheses test results. For $\mu_{\mathrm{LLM}}$ Vs $\mu_{\mathrm{human}}$ rows, mean values are presented and then shaded with a darker hue where $\mathrm{pvalue} > 0.05$. For $\mu_{1}$ - $\mu_{2}$ rows, the difference between means is presented and a darker shading highlights results with $\mathrm{pvalue} > 0.05$. 
    }
    \label{fig:pvalz-grid-all}
    \end{figure*}

%% file: pv-sari.tex
\begin{table}[htbp!] 
 \center
\begin{tabular}{lllcc}
\toprule
Model name & mean & ci (.95) & $n$\\
\midrule
\lblMISTRALflexi & 42.46 & 41.86 - 43.05 & 606 \\
\lblMISTRALstrict & 42.37 & 41.77 - 42.96 & 606 \\
\lblQWENflexi & 38.38 & 38.28 - 38.47 & 569 \\
\lblQWENstrict & 37.84 & 37.16 - 38.52 & 443 \\
\bottomrule
\end{tabular}
\caption{SARI simplification results}
\label{tbl:sari}
\end{table}

%% file: fig-corr-matrix.tex
 
\begin{figure*}
    \centering  
    \includegraphics[width=1.\textwidth]{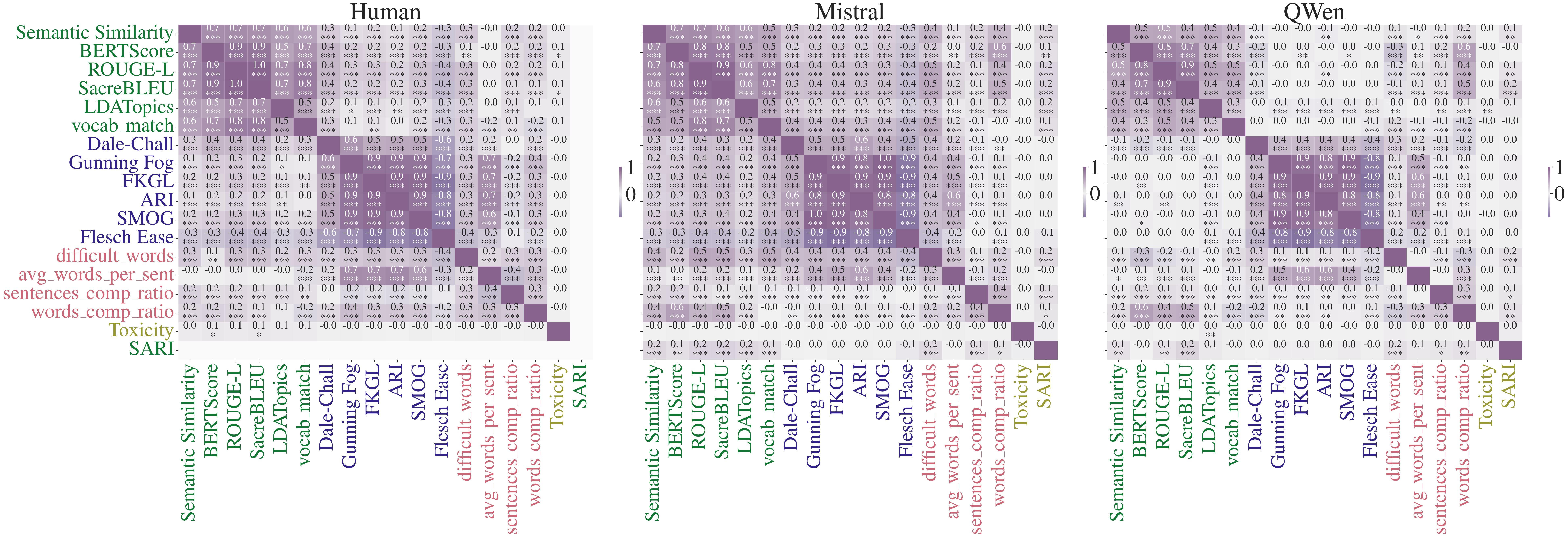}
    \caption{Correlations between metrics}
    \label{fig:corr-matrix-grid}
    \end{figure*}

%% file: fig-changes-min.tex
 
\begin{figure*}
    \centering  
    \begin{subfigure}[b]{1.\textwidth}
        \centering \raisebox{-\height}{\includegraphics[width=\linewidth]{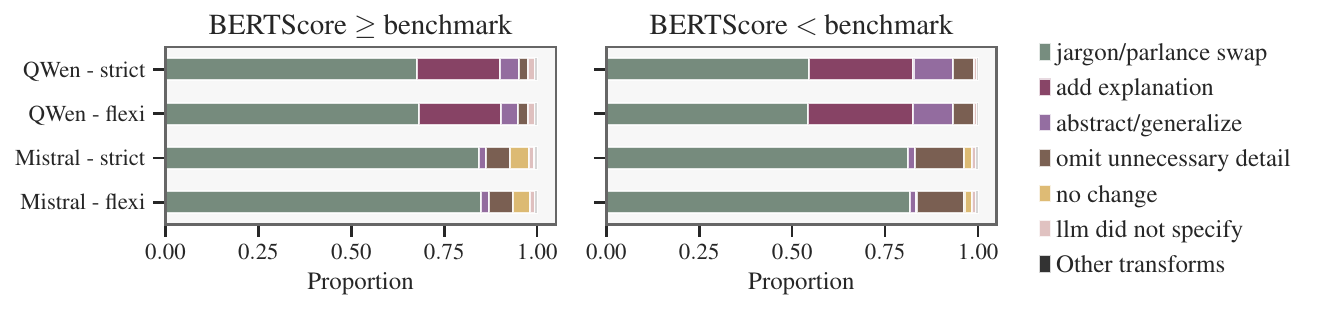}}
\label{fig:changes-BERTScore}
        \end{subfigure} 
    \caption{Self-reported rationale for changes made.}
    \label{fig:changes-pairs-mini}
    \end{figure*}

%% file: 04-a-discussion.tex

\paragraph{Overall performance.} 
The two general-purpose LLMs in this study demonstrate foundational readiness for biomedical text simplification with zero-shot prompting alone. 
Both models advance over prior SARI baselines, and Mistral's performance is competitive with GPT-4.1-mini. 
In addition, the models have statistically superior readability on four of six readability indices (SMOG Index, FKGL, ARI, and Flesch Reading Ease) and attain BERTScore values of $0.91$ and $0.89$ for~\MISTRAL~and~\QWEN, respectively.

\paragraph{Architectural differences.}
The instruction-tuned Mistral exhibits consistent performance across temperature settings, attaining accuracy scores that are statistically indistinguishable from human experts, and balancing readability and accuracy more effectively than QWen. 
Correlation analysis and self-reported rationales reveal Mistral has a conservative strategy that favors selective retention of specialized language, preserving semantic integrity.

Conversely, the reasoning-augmented QWen adopts a more exploratory approach to lexical complexity, achieving superior readability on four indices but risking semantic degradation (BERTScore 0.89, statistically below human score of 0.91). 
Its weaker coupling between readability and accuracy suggests disconnected optimization strategies by the model or a metric suite that does not adequately capture QWen's approach. 
Whether this observation points to a limitation (in model or metric suite) or a potential advantage  from exploration of a wider lexical search space warrants further investigation, particularly as LLMs incorporate more domain-specific training data.

\paragraph{Operational insights.}
The Dale-Chall and Gunning Fog indices, however, indicate statistically inferior results compared to the human benchmark. 
This discrepancy is traced to formulaic differences concerning the inherent definition (static word list) and weighting of lexical importance. 
Collectively, results point to syntactic mastery and lexical control emerging as the central challenge. 
Domain adaptation may benefit from integrating specialized lexical support rather than generalized knowledge expansion. 

Strong correlations among SMOG, FKGL, ARI, Flesh Reading Ease, and Gunning Fog ($\ge 0.7$) confirm redundancies within readability metrics, enabling principled metric reduction. 
Dale-Chall is distinctively different with coefficients in the range $[0.4 - 0.6]$.

While research recommends text expansion as a tactic for enhancing readability~\cite{cramm_best_2017,attal_dataset_2023}, the LLMs attain superior readability with more concise output, and text expansion serves to contain discourse integrity, additionally highlighting their efficiency potential.

%% file: 05-a-conclusion.tex

This study assesses the zero-shot text simplification capabilities of two general-purpose LLMs, both widely employed in research and practice, across a comprehensive set of readability and accuracy metrics. 
The instruction-tuned~\MISTRAL~exhibits superior operational robustness, attaining readability with a discourse fidelity comparable to human experts. Mistral displays a conservative algorithmic strategy that selectively balances lexical transformation and semantic preservation. 
In contrast, the reasoning-augmented~\QWEN~exhibits characteristic conceptual expansion attaining higher readability scores but risking semantic degradation. 
The models appear to have mastery over syntactic simplification, and the primary hurdle is lexical control.  
Additionally, the within- and cross-functional evaluation of various readability and accuracy metrics provides a basis for objective comparison and metric selection heuristics.

%% file: 04-b-limitations.tex

While we systematically assess two architectural classes using capable representatives, further evaluation across a wider range of LLMs is required to definitively characterize architectural differences in plain-text adaptation. 
In addition, despite employing established and transferable plain-text adaptation tactics, which should offer directional evidence or an informed starting point for broader application areas, the study explicitly addresses only one domain (biomedicine). 
Methodologically, the inherent limitations of conventional readability formulas must be recognized, and quantitative metrics should be complemented by human validation strategies as recommended by existing research~\cite{dai_critical_2024}.

%% file: 99-appendix.tex

\section{Methodological details}
\label{sec:appendix-methods}

\paragraph{Sample overview}
\Cref{tbl:sample-dessc} presents the sample sizes and request completion rates. 
A successful completion involves the transformation and return of a correctly formatted response for subsequent evaluation.
The instruction-tuned Mistral model maintains stable operational performance irrespective of the temperature configuration, attaining a completion rate of $85\%$ for the custom dataset.
Conversely, QWen exhibits sensitivity to temperature settings. 
All in all, reasonable sample sizes are obtained for subsequent interrogation.

\input{\TABLESDIR/pv-sample}

\paragraph{Metric properties}
\Cref{tbl:metric-propz} describes the metrics in the study. 
Traditional lexical alignment metrics like ROUGE and SacreBLEU, while not ideal for text simplification due to their reliance on lexical retention, serve as foundational measures for internal consistency and cross-study comparability.
Implementations are based on $evaluate$~\cite{noauthor_huggingface/evaluate_2025} and $textstats$~\cite{noauthor_textstat/textstat_2025} Python modules.

\paragraph{Regression by PCA analysis}
\label{sec:appendix-pca-method}
We set a metric as an independent variable and then ran a regression model on the first four PCA components of the other metrics as per \ref{eq:pca}.

	\begin{align*} 	
		y = & \mathbf{PCA}(\mathrm{StandardScaler} ( X ) ) \\
		\\
		\mathrm{metrics} \in \big\{ &\mathrm{BERTScore}, \\
		& \mathrm{Dale Chall}, \\
		& \mathrm{Flesch Ease}, \\
		& \mathrm{avg\_words\_per\_sent}, \\
		& \mathrm{vocab\_match}, \\
		& \mathrm{difficult\_words} \big\}
\label{eq:pca}
\end{align*}

\input{\FIGURESDIR/fig-pca-single}

\section{Prompt design}
\label{sec:appendix-prompt}
The full prompt and associated structured output validation models are presented in~\Cref{prompt-full},~\Cref{prompt-model}, and~\Cref{prompt-rationale}

\onecolumn

\input{\TABLESDIR/pv-metrics}

\begin{lstlisting}[style=promptstyle, caption={LLM prompt for text simplification}]
	SYSTEM PROMPT:		
	You're a scientific research assistant in the field of biomedical engineering, and  with excellent public dissemination and communication skills. Your task is to transform medical and scientific text, parlance and jargon into a version that is easy to read and understand for a layman with basic high school education. Relative to this task, this is what it means to simplify a text; it is to translate it from scientific parlance and jargon into lay easy to read language. 
		
	## INSTRUCTIONS
	**Transform the scientific text into a version that is easy to read and understand for a layman**
	
	1. **Your transformation operations work at a sentence level**. 
	- You must operate at a sentence level. 
	- For instance, a title text is already a sentence, while an abstract or a paragraph of text is not. A paragraph of text has multiple sentences, so you **MUST split paragraphs into a list of sentences first**, and index each sentence starting from `0` like one would a python list. 
	- Therefore, if the input is a single sentence, operate on it directly. Else if it is a paragraph, first split it into its constituent sentences, then operate on each sentence at a time. 
	- If the context a sentence belongs to is provided, consider that context only as a guide to help you better transform the sentence in question.
	- That also means you cannot summarize a paragraph as that risks loss of meaning and information. **Each sentence in the original text must be accounted for**.  
		
	2. **For each sentence, consider the following possible transformations** that might realise simpler sentences, that are easy to read and understand for a layman. 
	- You may split a sentence into 2 or more sentences as part of the simplification transform. For instance, in the case of long complex sentences
	- Identify medical and scientific parlance and jargon and substitute it with equivalent lay terms 
	- Where equivalent lay terms are not available, consider providing an explanation or clarifying examples 
	- If the scientific terms and phrases are too granular and a more generic/abstract form would suffice for a layman, and without loss of meaning, prefer the more general/abstract form/terms
	- You may omit (transform entire sentence into an empty string) sentences that are intended for a scientific audience and whose removal does not alter the understanding of the meaning of the broader document text
	- Prefer active over passive voice. 
	- Add subheadings at the start of a sentences. For instance, "Aim:" can be added to an objective sentence, while "Results:" may be added to a findings/results sentence. 
	- When providing quantitative results, remove scientific detaills such confidence intervals (CI), p-values, etc  
		
	Other guidelines 
	- **STRICTLY, DO NOT summarize the research or extract and transform texts that are not explicitly requested for in the input**
	- **STRICTLY, DO NOT synthesize or summarize the text.** 
	- **Your task and goals is to traslate from medical/scientific parlance and jargon into layman easy to read and understand language**.
					
	## OTHER GUIDELINES TO OBSERVE
	- Be concise and succinct
	- if you don't know don't not create/imagine facts
	- use the string `**N/A**` for empty non-numeric values  and the number `-99` for empty numeric values if you must report a value and yet you don't know.
	- Striclty adhere to the output format requested. Only extract and return as per the output format or JSON schema; no other detail!! 
	- If JSON output is requested, return only valid json output AND as per the indicated format/schema. **Strictly return valid JSON output and as per indicated output JSON schema**, AND enclose the JSON output in a code block e.g "```json <only requested output as per output JSON schema goes here>```".

	USER PROMPT: 
	## Text to transform
	<input> {input_data} </input>
		
	OUTPUT FORMAT:
	{Output format representation of the Pydantic object}

\label{prompt-full}
\end{lstlisting}  
\twocolumn

\begin{lstlisting}[style=pythonstyle, caption={Pydantic model for structured LLM output}] 
	class PlabaSentence(BaseModel):
		index_of_input_sentence:Optional[int] = Field(None, description="A. The index of the sentence in the `input`/<input> text if the `input`/<input> text is a paragraph OR B. The index of a sentence in the `context`/<context> paragraph if the `input`/<input> text is a single sentence. The index of a sentence is its sequence number in the containing paragraph, and the first sentence is indexed 0, the sencond 1 and so on.")
		input_sentence:Optional[str] = Field(None, description="The original sentence, as provided by user in `input`/<input>, and which is to be simplified/transformed.")
		output_sentence:Optional[str] = Field(None, description="The new transformed/simplified sentence. It is okay to return the original sentence here if and only if the sentence does not need changing/simplification. This can also be an empty string if simplication entails complete omission of the sentence.")
		changes_made:Optional[List[SimplificationChange]] = Field(None, description="Strict list (at most 10) of distinct changes/transformations made to this sentence, if at all.")
		rationale:Optional[str] = Field(None, description="Brief summary, in under 30 words, of the rationale behind transforming that sentence in that manner.")
	
	class PlabaDoc(BaseModel):
		list_of_requested_responses:Optional[List[PlabaSentence]] = Field(None, description="The list of simplified sentences. Each sentence in the `input`/<input> must be accounted for accordingly.")	
		
\label{prompt-model}
\end{lstlisting}

\begin{lstlisting}[style=pythonstyle, caption={Pydantic model for capturing the rationale behind LLM response}] 	
	class SimplificationTransforms(Enum):
		SWAP_OUT_JARGON = "jargon/parlance swap"
		EXPLAIN_JARGON_OR_PHRASE = "explain jargon"
		SWAP_GRANULAR_DETAILS_FOR_GENERAL = "abstract granular"
		OMIT_UNECESSARY_SCI_DETAILS = "omit unnecessary"
		NO_CHANGE_NECESSARY = "no change necessary" 
		OTHER = "other"
			
	class SimplificationChange(BaseModel):
		input_phrase:Optional[str] = Field(None,description="The original word or phrase as provided in the `input`/<input> text by the user.")
		update_phrase:Optional[str] = Field(None,description="The new/transformed version.") 
		update_type:Optional[SimplificationTransforms] = Field(None,description="kind of transformation/change it is.")
		update_rationale:Optional[str] = Field(None,description="Brief explanation, in under 30 words, for this particular change.")

\label{prompt-rationale} 		
\end{lstlisting}


\section{Additional results}
\label{sec:appendix-results}

\paragraph{Github page with analysis results} 
\href{\GITPage}{Github page with comprehensive analysis output (CSVs) and reference tables.}

\input{\FIGURESDIR/fig-mu-distz-all}

\input{\FIGURESDIR/fig-corro-slxn}
%
%

\input{\FIGURESDIR/fig-changes}

%% file: pv-sample.tex
\begin{table} 
 \center
\begin{tabular}{llccc}
\toprule
 & Simplification & $n$ & Completion & $n$ \\
Dataset & Model & Documents & Rate & Evaluations \\
\midrule
Benchmark & human & 748 & 1.00 & 26,926 \\
Control & \lblMISTRALflexi & 606 & 0.81 & 15,322 \\
Control & \lblMISTRALstrict & 606 & 0.81 & 15,288 \\
Control & \lblQWENflexi & 569 & 0.76 & 13,533 \\
Control & \lblQWENstrict & 443 & 0.59 & 11,182 \\
Custom & \lblMISTRALflexi & 3,218 & 0.85 & 78,030 \\
Custom & \lblMISTRALstrict & 3,217 & 0.85 & 77,994 \\
Custom & \lblQWENflexi & 3,672 & 0.97 & 69,336 \\
Custom & \lblQWENstrict & 2,453 & 0.65 & 59,340 \\
\bottomrule
\end{tabular}
\caption{Sample overview.}
\label{tbl:sample-dessc}
\end{table}

%% file: fig-pca-single.tex
 
\begin{figure}[htbp!]  
    \centering  
    \includegraphics[width=0.98\linewidth]{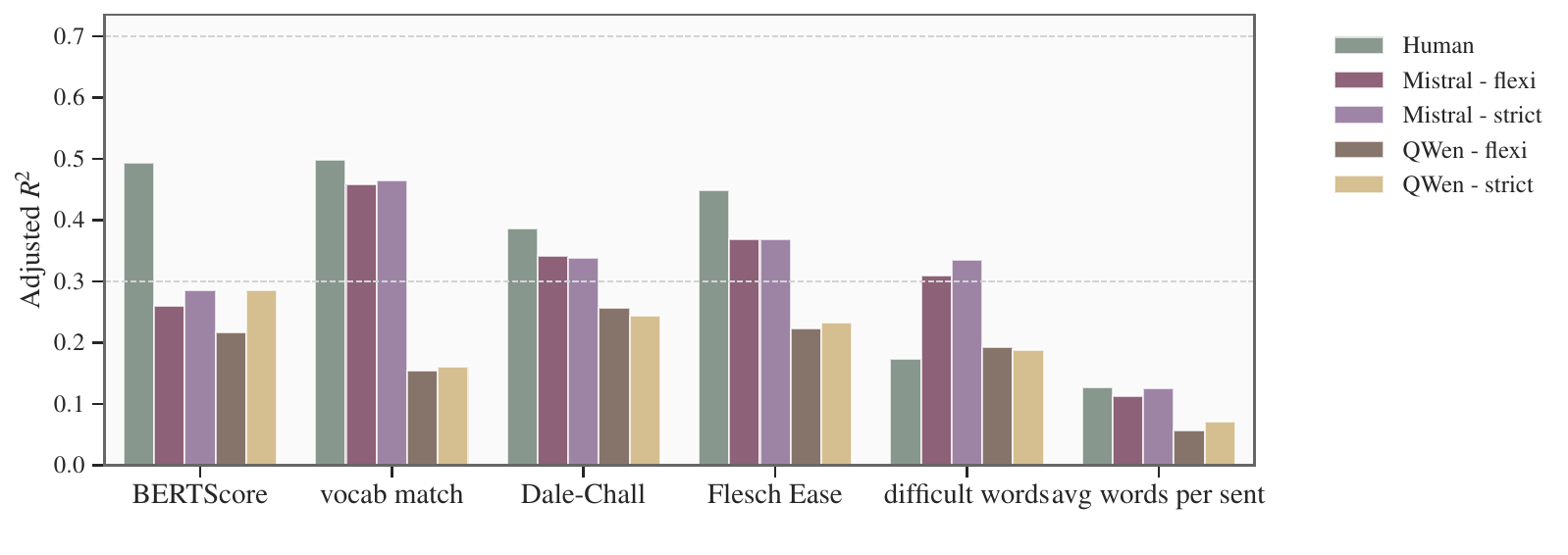} 
    \caption{Regression by PCA results}
    \label{fig:pca-reg-bars-yi}
\end{figure}

%% file: pv-metrics.tex
\renewcommand{\arraystretch}{1.2}\begin{table}[hb!]
\centering
\begin{tabular}{w{l}{4.8cm}>{\small}w{l}{4.15cm}>{\small}p{\dimexpr\textwidth-9.0cm\relax}}
\toprule
A. Foundational/supplementary metrics &  & Computation notes \\
\midrule 
n\ words &  & $\mathbf{W}$ \\
n\ sentences &   & $\mathbf{S}$ \\
n\ syllables\ in\ word &  & $\mathbf{P}$  ;   Typically polysyllabic if $ \mathbf{P}\ge 3$ \\
avg\ words\ per\ sent &   & $\mathbf{L} = \frac{ \mathbf{W} }{ \mathbf{S} }$ \\
difficult\ words &  & $  \mathbf{V} = \frac{ \sum \mathbb{I}( ( w \notin \mathrm{DaleChallList}) |  (\mathbf{P} \ge 3))  }{ \mathbf{W} }  * 100$ \\
sentences\ comp\ ratio & Compression or expansion ratio & $\frac{\mathbf{S} {\mathrm{simplified}}}{\mathbf{S} {\mathrm{source}}}$ \\
words\ comp\ ratio & Compression or expansion ratio & $\frac{\mathbf{W} {\mathrm{simplified}}}{\mathbf{W} {\mathrm{source}}}$ \\
vocab\ match & Terms (lemmatized) & $\mathrm{Jaccard} ( \mathbf{T} {\mathrm{simplified}} ,   \mathbf{T} {\mathrm{source}} )$ \\

Toxicity & Content safety & Roberta-hate-speech-dynabench-r4 \\
\cmidrule(r){1-3}
\multicolumn{3}{l}{ B. Discourse fidelity metrics}\\
\cmidrule(r){1-3}
Semantic Similarity & QWen2.5 32B Embeddings & $\cos ( \mathbf{Doc} {\mathrm{simplified}}, \mathbf{Doc} {\mathrm{source}} )$ \\
BERTScore~\citep{zhangbertscore2020} &   N-gram-based & F1 score value. (Roberta Large) \\
ROUGE-L~\citep{linrouge2004} &  N-gram-based &  Longest common subsequence. With stemming. \\
SacreBLEU~\citep{postcall2018} &   N-gram-based & Defaults \\
LDATopics &  Terms & $\mathrm{Jaccard} ( \mathbf{T} {\mathrm{simplified}} ,   \mathbf{T} {\mathrm{source}} )$ \\

\cmidrule(r){1-3}
\multicolumn{3}{l}{ C. Readability metrics}\\
\cmidrule(r){1-3}
SARI~\citep{xuoptimizing2016} & System goodness, n-gram based & $ \frac{ \mathbf{F1} {add}  + \mathbf{F1} {keep} + \mathbf{Pr} {del} } { 3 }$ ;  $\mathbf{F1}$ score, $\mathbf{Pr}$ecision score \\
SMOG~\citep{smog} & USA School Grade & $1.0430  *   \sqrt{  (\sum {w} \mathbb{I}(\mathbf{P} \ge 3)  * \frac{30}{\mathbf{S}}   }  )  + 3.1291  $ \\
Gunning Fog~\citep{gunningtechniquenodate} & USA School Grade & $0.4 * (  \mathbf{V}   +    \mathbf{L}  )$ \\
ARI~\citep{smithautomated1967} & USA School Grade & $ (4.71 * \frac{\mathrm{nCharacters}}{\mathbf{W}}) +  (0.5 * \mathbf{L})  - 21.43$ \\
Dale-Chall & USA School Grade & $ (0.1579 * \frac{  \sum \mathbb{I}( w \notin \mathrm{DaleChallList})    }{\mathbf{W}} * 100)   +  (0.0496 * \mathbf{L})  \space [+ 3.6365 ] $ \\
FKGL~\citep{kincaidelectronic1988} & USA School Grade & $-15.59  +  (11.8 * \frac{\sum {w}(\mathbf{P})}{\mathbf{W}})    + (0.39 *  \mathbf{L})   $ \\
Flesch Ease~\citep{klareautomation1969} &  & $206.835 - (84.6 * \frac{\sum {w}(\mathbf{P})}{\mathbf{W}} )   - (1.015 *  \mathbf{L} )  $ \\

\bottomrule
\end{tabular}
\caption{The suite of metrics in the evaluation.}
\label{tbl:metric-propz}
\end{table}
\renewcommand{\arraystretch}{1.0}

%% file: fig-mu-distz-all.tex
 
\begin{figure}[htbp!]  
    \centering   
\begin{subfigure}[b]{0.49\linewidth}
        \centering \raisebox{-\height}{\includegraphics[width=\linewidth, height=4cm]{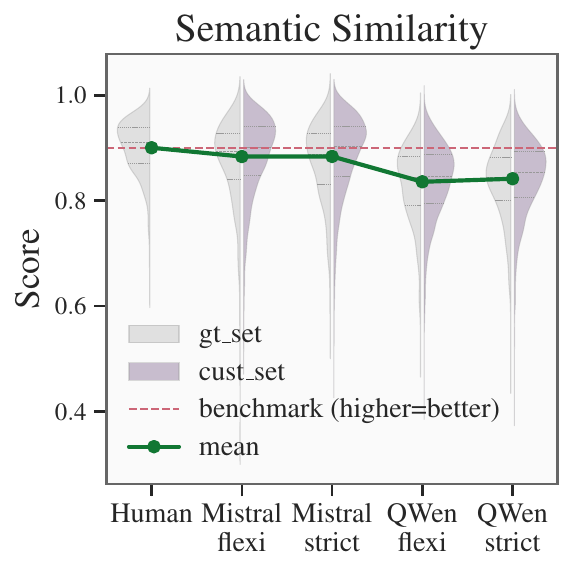}}
\label{fig:mean-distz-semantic-similarity-doc}
        \end{subfigure}
 \begin{subfigure}[b]{0.49\linewidth}
        \centering \raisebox{-\height}{\includegraphics[width=\linewidth, height=4cm]{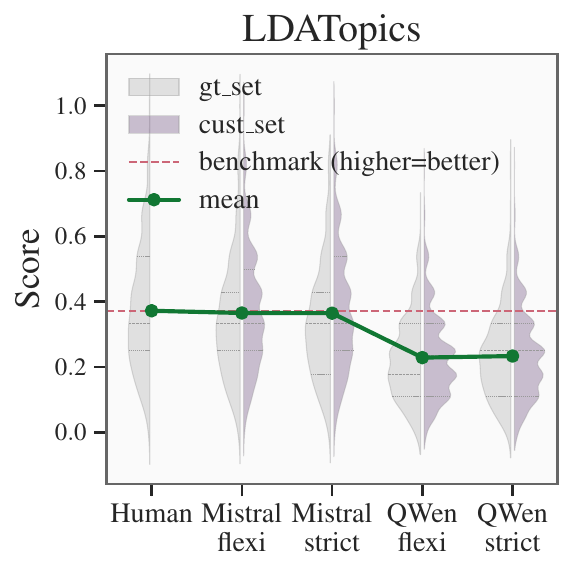}}
        \label{fig:mean-distz-lda-topics}
        \end{subfigure} 
\begin{subfigure}[b]{0.49\linewidth}
        \centering \raisebox{-\height}{\includegraphics[width=\linewidth, height=4cm]{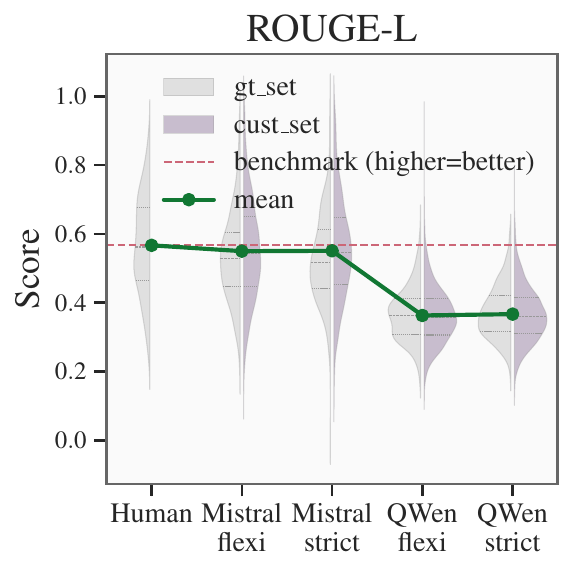}}
\label{fig:mean-distz-ROUGE}
        \end{subfigure}
\begin{subfigure}[b]{0.49\linewidth}
        \centering \raisebox{-\height}{\includegraphics[width=\linewidth, height=4cm]{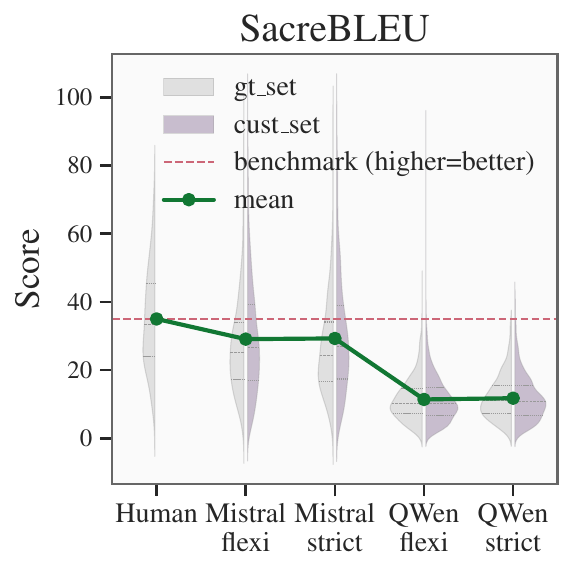}}
\label{fig:mean-distz-SacreBLEU}
        \end{subfigure}\hfill
\begin{subfigure}[b]{0.49\linewidth}
    \centering \raisebox{-\height}{\includegraphics[width=\linewidth, height=4cm]{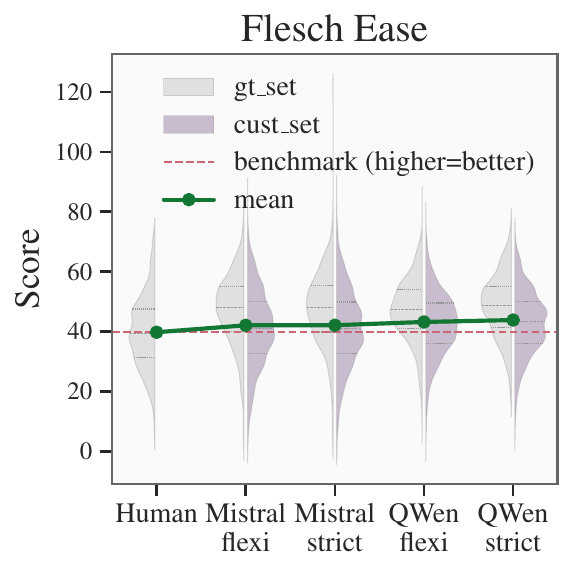}}
    \label{fig:mean-distz-readability-flesch-ease}
    \end{subfigure}
\begin{subfigure}[b]{0.49\linewidth}
        \centering \raisebox{-\height}{\includegraphics[width=\linewidth, height=4cm]{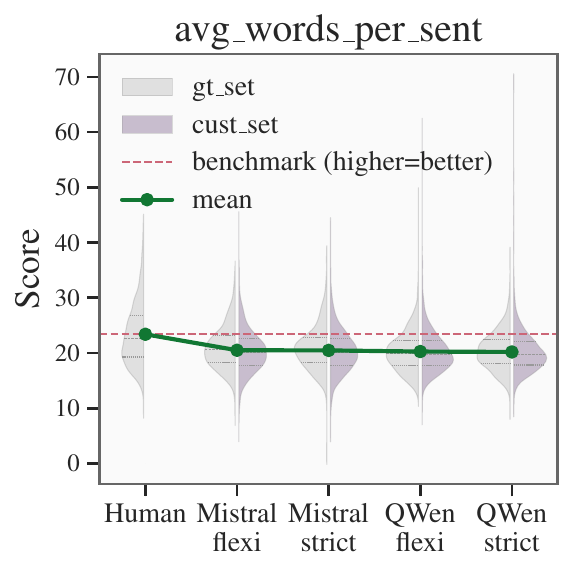}}
\label{fig:mean-distz-avg-words-per-sent}
        \end{subfigure}
\begin{subfigure}[b]{0.49\linewidth}
        \centering \raisebox{-\height}{\includegraphics[width=\linewidth, height=4cm]{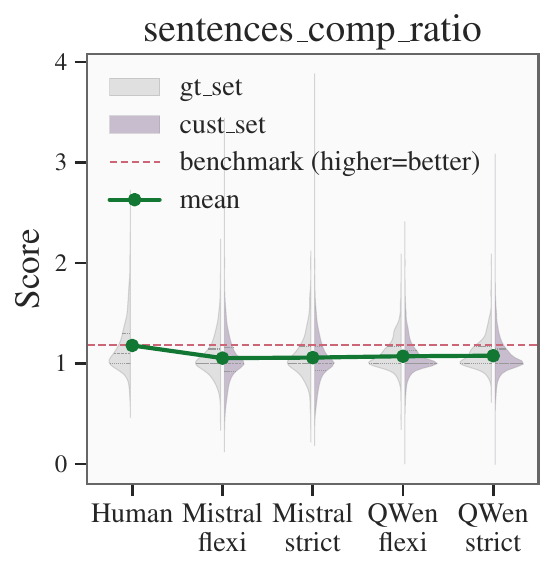}}
\label{fig:mean-distz-sentences-comp-ratio}
        \end{subfigure}
\begin{subfigure}[b]{0.49\linewidth}
        \centering \raisebox{-\height}{\includegraphics[width=\linewidth, height=4cm]{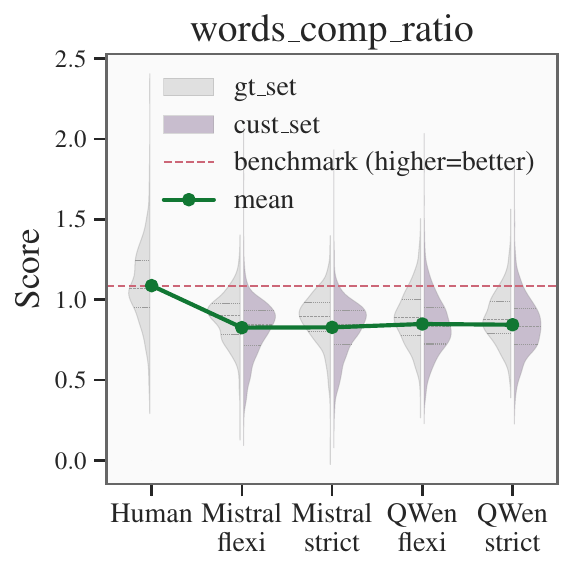}}
\label{fig:mean-distz-words-comp-ratio}
        \end{subfigure}\hfill
\begin{subfigure}[b]{0.49\linewidth}
        \centering \raisebox{-\height}{\includegraphics[width=\linewidth, height=4cm]{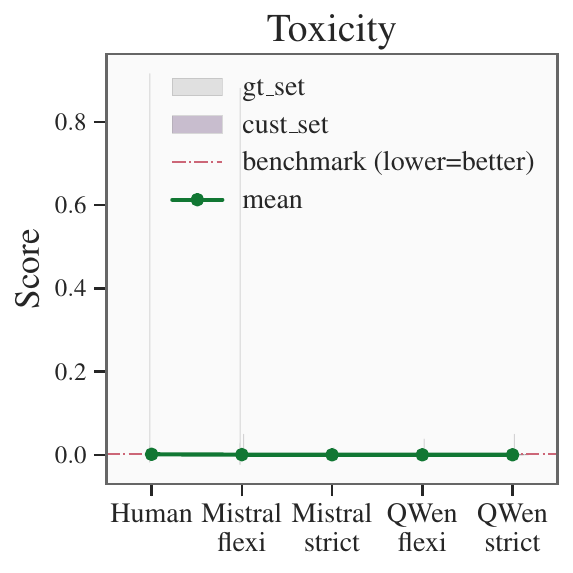}}
\label{fig:mean-distz-toxicity}
        \end{subfigure}
\begin{subfigure}[b]{0.49\linewidth}
        \centering \raisebox{-\height}{\includegraphics[width=\linewidth, height=4cm]{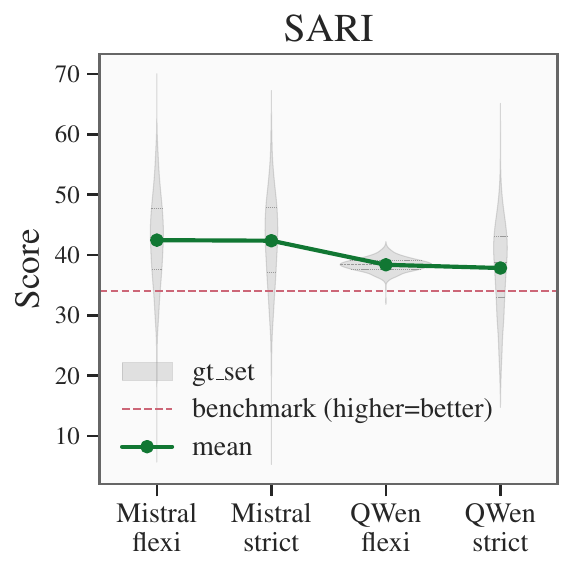}}
\label{fig:mean-distz-SARI}
        \end{subfigure}
    \caption{Average performance and underlying data distributions (for other metrics)}
    \label{fig:means-distz-all-tog}
    \end{figure}

%% file: fig-corro-slxn.tex
 
\begin{figure}[htbp!]  
    \centering  
    \begin{subfigure}[b]{0.49\linewidth}
        \centering \raisebox{-\height}{\includegraphics[width=\linewidth]{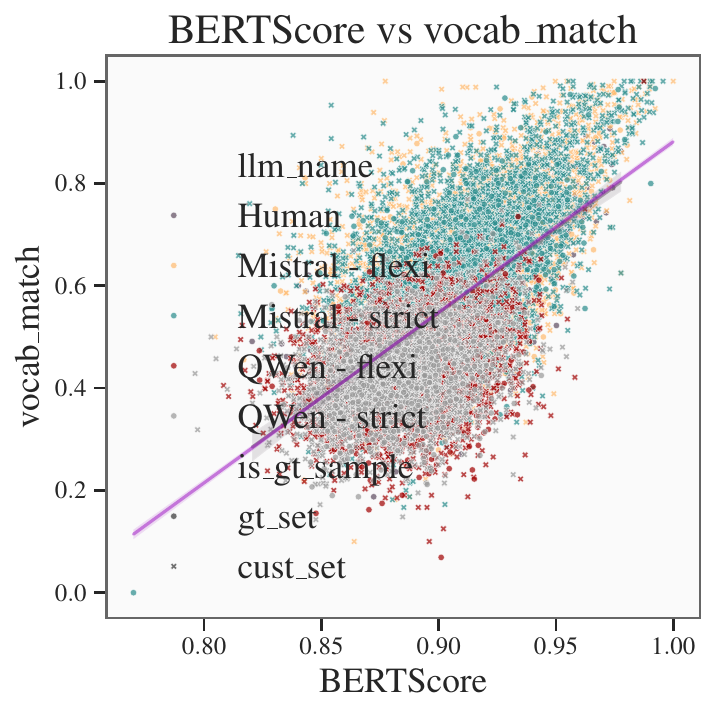}}
\label{fig:corro-slxn-bert}
        \end{subfigure}
\begin{subfigure}[b]{0.49\linewidth}
        \centering \raisebox{-\height}{\includegraphics[width=\linewidth]{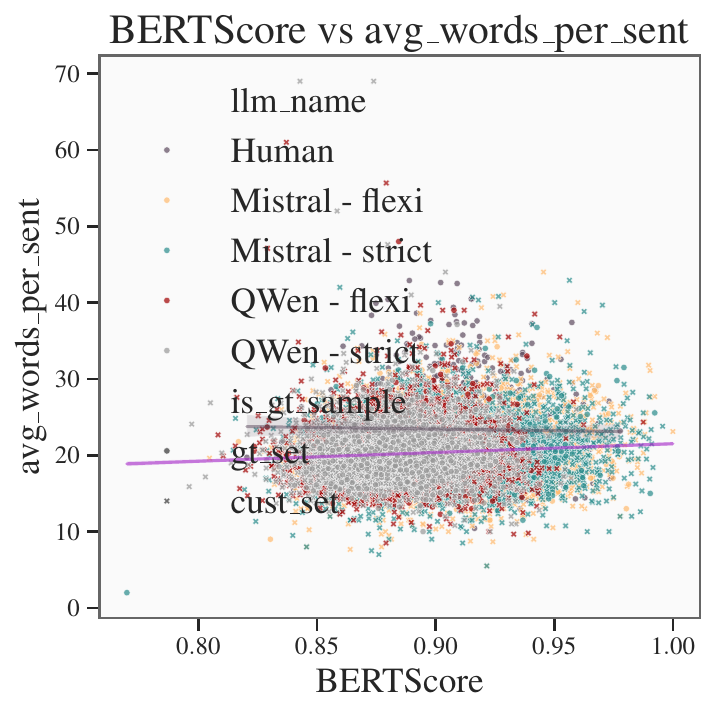}}
\label{fig:corro-slxn-bert}
        \end{subfigure}
\begin{subfigure}[b]{0.49\linewidth}
        \centering \raisebox{-\height}{\includegraphics[width=\linewidth]{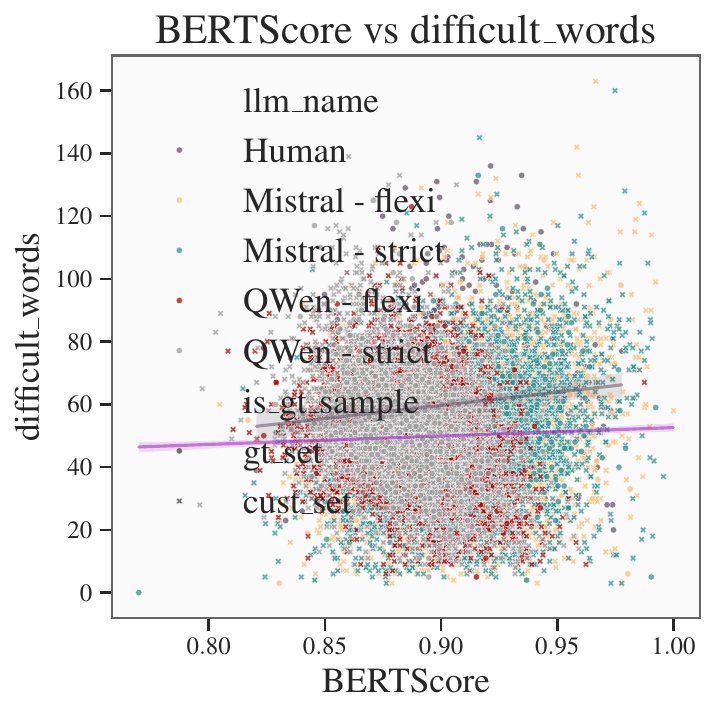}}
\label{fig:corro-slxn-bert}
        \end{subfigure}
\begin{subfigure}[b]{0.49\linewidth}
        \centering \raisebox{-\height}{\includegraphics[width=\linewidth]{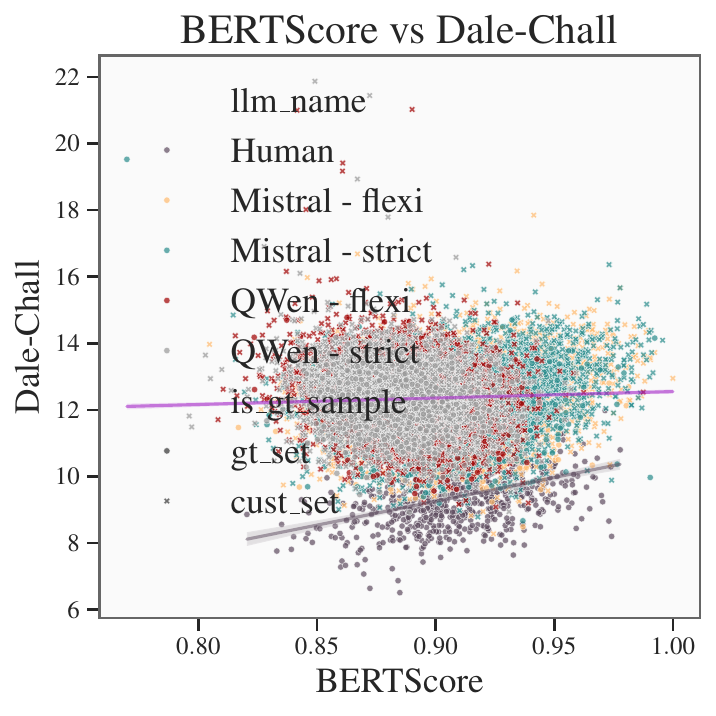}}
\label{fig:corro-slxn-bert}
        \end{subfigure}
\begin{subfigure}[b]{0.49\linewidth}
        \centering \raisebox{-\height}{\includegraphics[width=\linewidth]{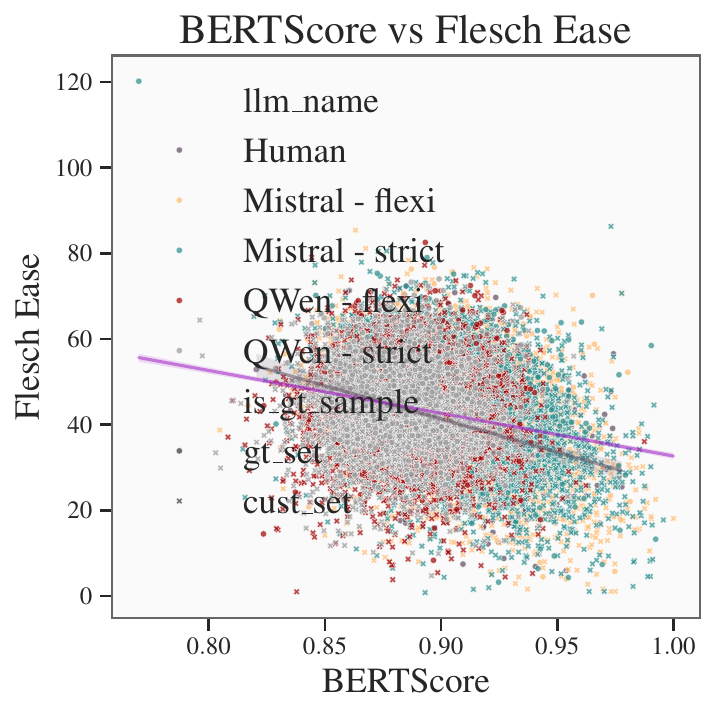}}
\label{fig:corro-slxn-bert}
        \end{subfigure}
\hfill\begin{subfigure}[b]{0.49\linewidth}
        \centering \raisebox{-\height}{\includegraphics[width=\linewidth]{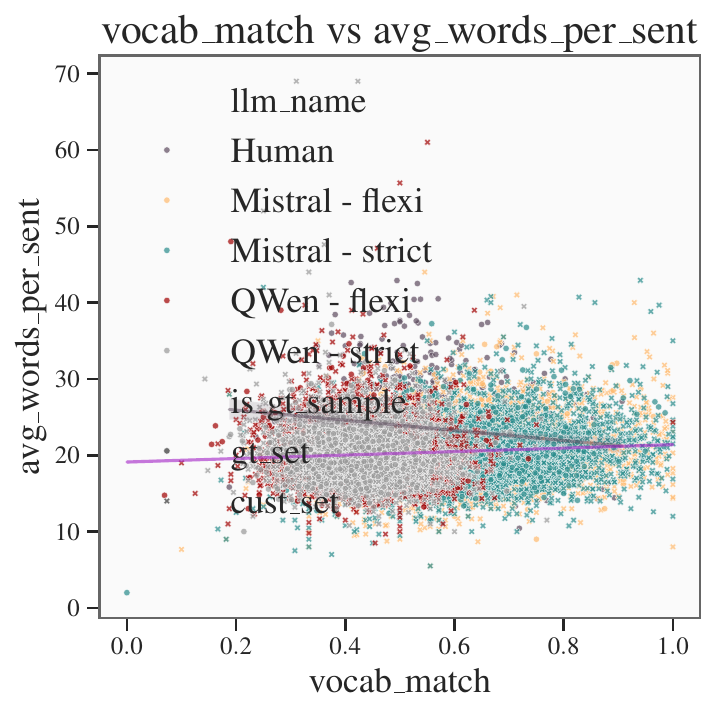}}
\label{fig:corro-slxn-vocab}
        \end{subfigure}
\begin{subfigure}[b]{0.49\linewidth}
        \centering \raisebox{-\height}{\includegraphics[width=\linewidth]{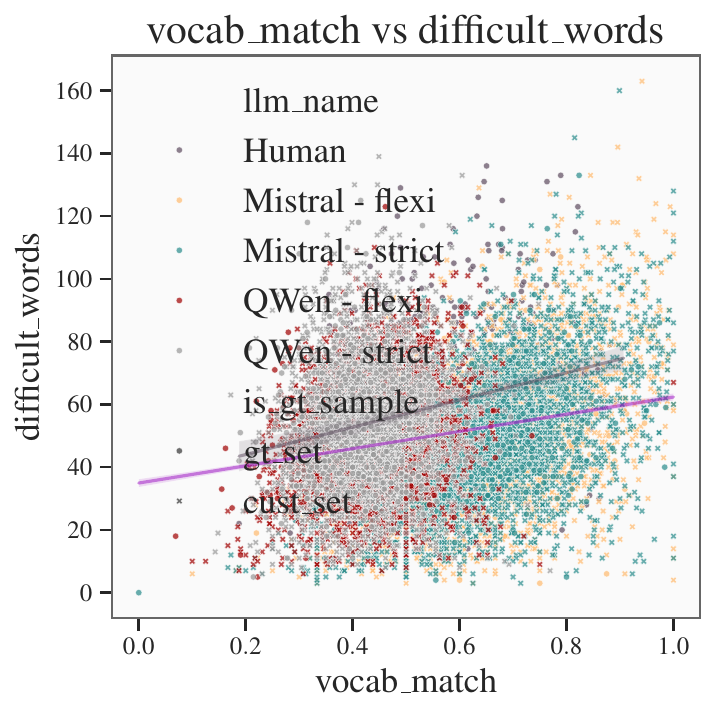}}
\label{fig:corro-slxn-vocab}
        \end{subfigure}
\begin{subfigure}[b]{0.49\linewidth}
        \centering \raisebox{-\height}{\includegraphics[width=\linewidth]{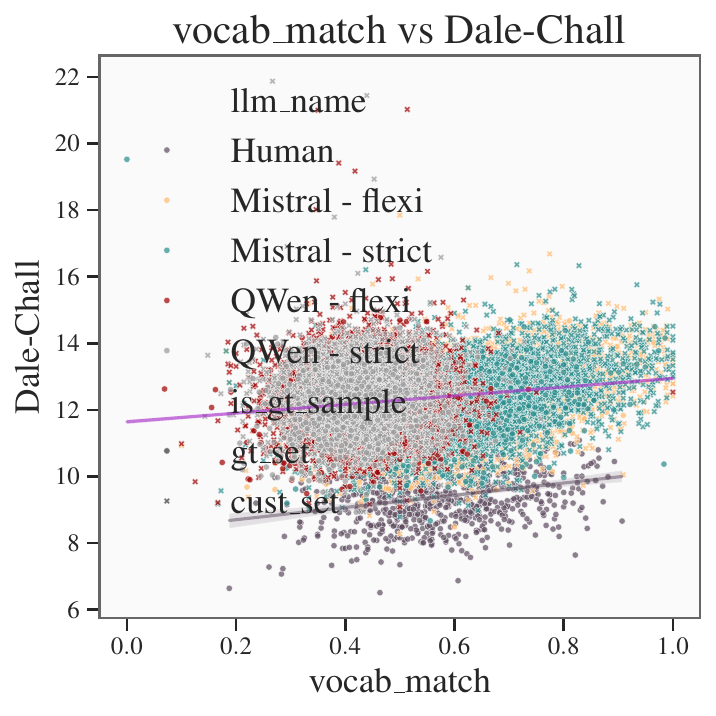}}
\label{fig:corro-slxn-vocab}
        \end{subfigure}
\begin{subfigure}[b]{0.49\linewidth}
        \centering \raisebox{-\height}{\includegraphics[width=\linewidth]{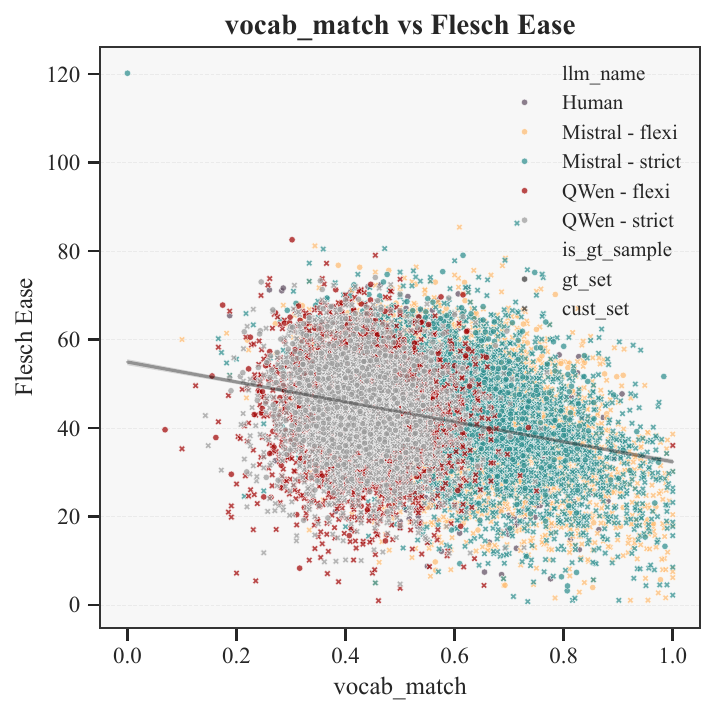}}
\label{fig:corro-slxn-vocab}
        \end{subfigure}

    \caption{Example pairwise correlation plots}
    \label{fig:corro-slxn-all}
    \end{figure}
    

%% file: fig-changes.tex
 
\begin{figure*}[htbp!]  
    \centering   
\begin{subfigure}[b]{0.99\linewidth}
        \centering \raisebox{-\height}{\includegraphics[width=\linewidth]{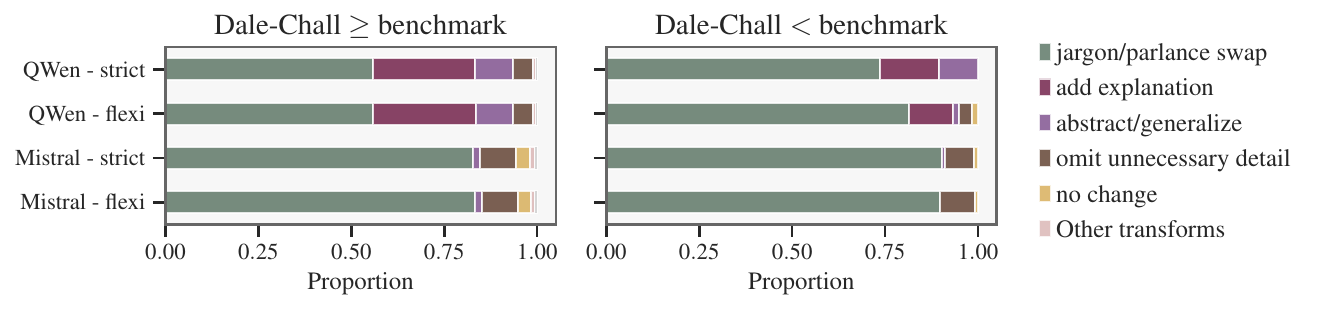}}
\hfill\label{fig:changes-dale-chall}
        \end{subfigure}
        \begin{subfigure}[b]{0.99\linewidth}
        \centering \raisebox{-\height}{\includegraphics[width=\linewidth]{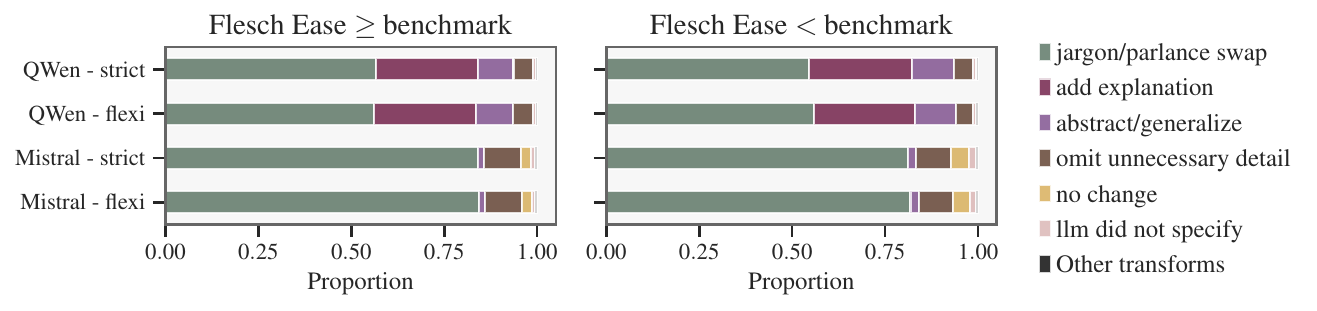}}
        \label{fig:changes-readability-flesch-ease}
        \end{subfigure}
\hfill\begin{subfigure}[b]{0.99\linewidth}
        \centering \raisebox{-\height}{\includegraphics[width=\linewidth]{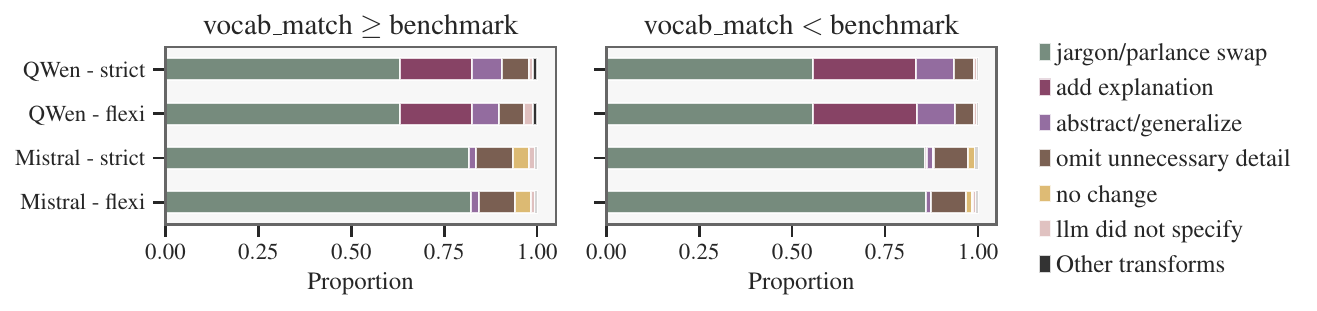}}
\label{fig:changes-vocab-match}
        \end{subfigure}
\begin{subfigure}[b]{0.99\linewidth}
        \centering \raisebox{-\height}{\includegraphics[width=\linewidth]{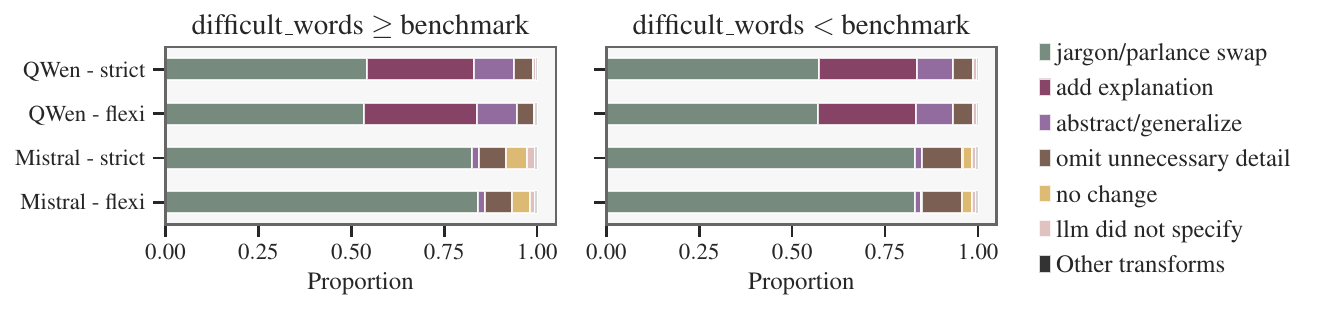}}
\label{fig:changes-difficult words}
        \end{subfigure}
\hfill
\begin{subfigure}[b]{0.69\linewidth}
	\centering \raisebox{-\height}{\includegraphics[width=\linewidth]{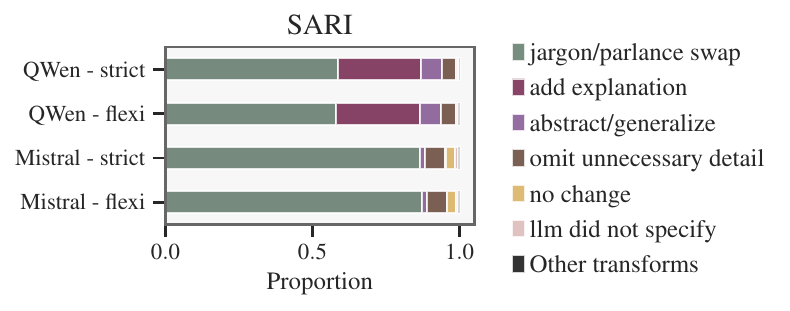}}
	\label{fig:changes-sari}
\end{subfigure}
    \caption{LLM self-reported rationale for text simplification changes made.}
    \label{fig:changes-pairs-selection}
    \end{figure*}